\documentclass[11pt,a4paper]{article}
\usepackage[hyperref]{acl2021}
\usepackage{times}
\usepackage{latexsym}

\usepackage{microtype}
\usepackage{ amssymb }
\usepackage{amsmath}
\usepackage{booktabs}
\usepackage{multirow}
\usepackage{algorithm}
\usepackage{caption}
\usepackage{algorithmic}

\allowdisplaybreaks[4]

\usepackage{bbm}
\usepackage{bm}

\makeatletter
\newcommand{\thickhline}{%
    \noalign {\ifnum 0=`}\fi \hrule height 2pt
    \futurelet \reserved@a \@xhline
}
\makeatother
\usepackage{tabularx}
\usepackage{adjustbox}
\usepackage{soul}
\usepackage{caption}
\usepackage{subcaption}
\usepackage[normalem]{ulem}
\usepackage{xr}

\aclfinalcopy % Uncomment this line for the final submission
\title{A Mutual Information Maximization Approach for \\
the Spurious Solution Problem in Weakly Supervised Question Answering}

\author{
    Zhihong Shao\textsuperscript{1}, Lifeng Shang\textsuperscript{2}, Qun Liu\textsuperscript{2}, Minlie Huang\textsuperscript{1}\thanks{*Corresponding author: Minlie Huang.}\\
    \textsuperscript{1}The CoAI group, DCST, Tsinghua University, Institute for Artificial Intelligence; \\
    \textsuperscript{1}State Key Lab of Intelligent Technology and Systems; \\
    \textsuperscript{1}Beijing National Research Center for Information Science and Technology; \\
    \textsuperscript{1}Tsinghua University, Beijing 100084, China \\
    \textsuperscript{2}Huawei Noah’s Ark Lab \\
    {\tt szh19@mails.tsinghua.edu.cn, aihuang@tsinghua.edu.cn} \\
    {\tt \{shang.lifeng, qun.liu\}@huawei.com}}

\date{}

\begin{document}
\maketitle
\begin{abstract}
Weakly supervised question answering usually has only the final answers as supervision signals while the correct solutions to derive the answers are not provided. This setting gives rise to the \textit{spurious solution problem}: there may exist many spurious solutions that coincidentally derive the correct answer, but training on such solutions can hurt model performance (e.g., producing wrong solutions or answers). For example, for discrete reasoning tasks as on DROP, there may exist many equations to derive a numeric answer, and typically only one of them is correct. Previous learning methods mostly filter out spurious solutions with heuristics or using model confidence, but do not explicitly exploit the semantic correlations between a question and its solution.
In this paper, to alleviate the spurious solution problem, we propose to explicitly exploit such semantic correlations by maximizing the mutual information between question-answer pairs and predicted solutions. Extensive experiments on four question answering datasets show that our method significantly outperforms previous learning methods in terms of task performance and is more effective in training models to produce correct solutions.
% and obtains new state-of-the-art results on WikiSQL under the weakly supervised setting.
\end{abstract}

\section{Introduction}
\begin{figure}[!t]
    \centering
    \includegraphics[width=0.45\textwidth]{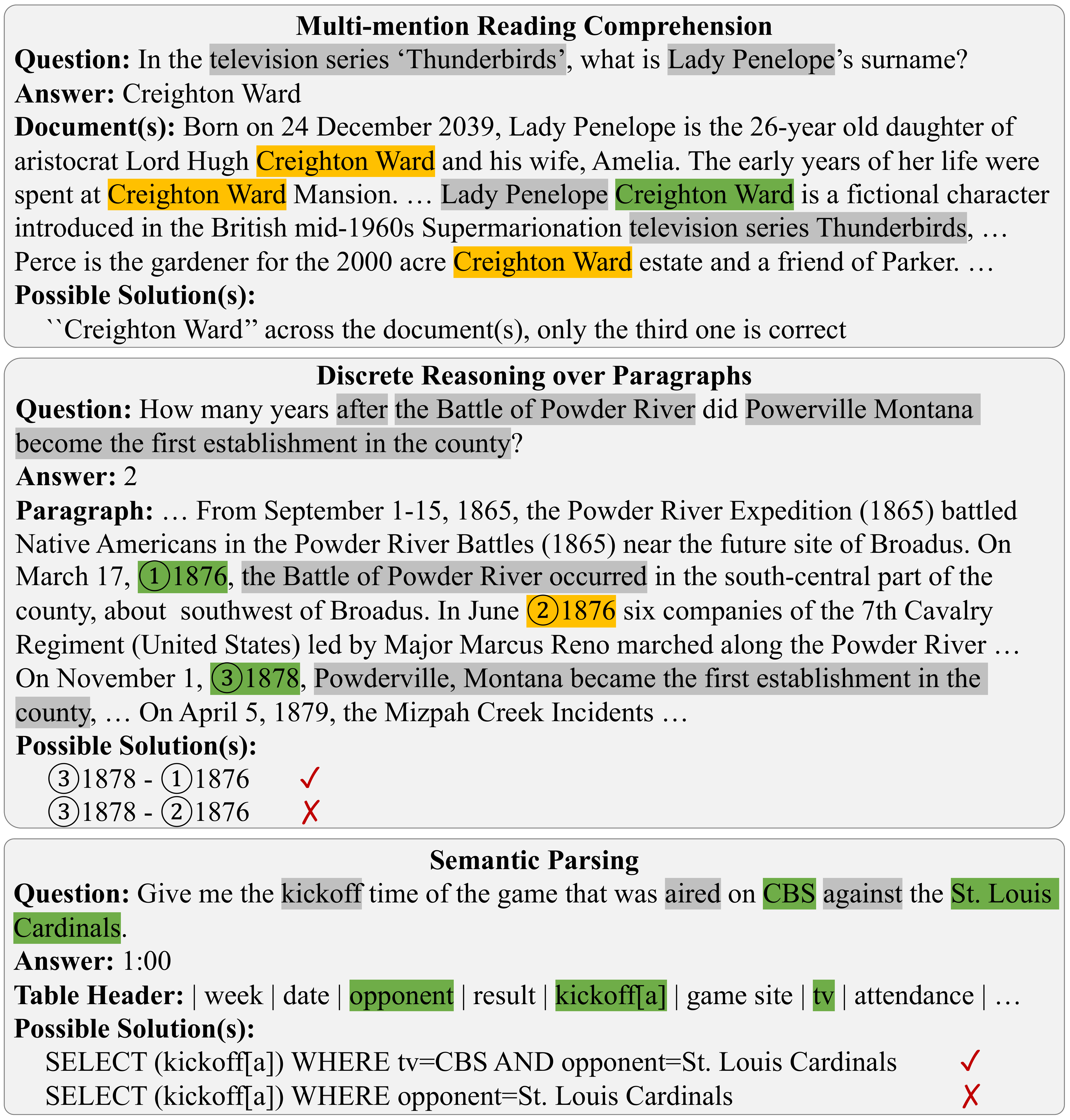}
    \caption{Examples from three weakly supervised QA tasks, i.e., multi-mention reading comprehension, discrete reasoning, and semantic parsing. Spans in dark gray and green denote semantic correlations between a question and its solution, while spans in orange are spurious information and should not be used in a solution.}
    \label{fig:intro_case}
\end{figure}
Weakly supervised question answering is a common setting of question answering (QA) where only final answers are provided as supervision signals while the correct solutions to derive them are not. This setting simplifies data collection, but exposes model learning to the \textit{spurious solution problem}: there may exist many spurious ways to derive the correct answer, and training a model with spurious solutions can hurt model performance (e.g., misleading the model to produce unreasonable solutions or wrong answers). As shown in Fig \ref{fig:intro_case}, for multi-mention reading comprehension, many mentions of an answer in the document(s) are irrelevant to the question; for discrete reasoning tasks or text2SQL tasks, an answer can be produced by the equations or SQL queries that do not correctly match the question in logic. 
% It is important but also challenging to solve the spurious solution problem, as a model learns to understand and answer a question if trained to produce correct solutions, but recognizing correct solutions itself needs understanding the question and the logic of solutions.
% Solving the spurious solution problem is challenging, as it is usually unknown in advance how to evaluate the consistency between a question and a solution.
% It is challenging to solve the spurious solution problem, as how to evaluate the correctness of a solution is usually not known in advance.

Some previous works heuristically selected one possible solution per question for training, e.g., the first answer span in the document \cite{DBLP:conf/acl/JoshiCWZ17,DBLP:conf/nips/TayLHS18,DBLP:conf/acl/TalmorB19}; some treated all possible solutions equally and maximized the sum of their likelihood (maximum marginal likelihood, or MML) \cite{DBLP:conf/iclr/SwayamdiptaPK18,DBLP:conf/acl/GardnerC18,DBLP:conf/acl/LeeCT19}; many others selected solutions according to model confidence \cite{DBLP:conf/nips/LiangNBLL18,DBLP:conf/emnlp/MinCHZ19}, i.e., the likelihood of the solutions being derived by the model.
% , which may correlate with many factors other than logical correctness, e.g., the length of the solutions. 
A drawback of these methods is that they do not explicitly consider the mutual semantic correlations between a question and its solution when selecting solutions for training.

Intuitively speaking, a question often contains vital clues about how to derive the answer, and a wrong solution together with its context often fails to align well with the question. Take the discrete reasoning case in Fig \ref{fig:intro_case} as an example. To answer the question, we need to know the start year of \textit{the Battle of Powder River}, which is answered by the first \textit{1876}; the second \textit{1876} is irrelevant as it is the year of an event that happened during the battle. 

To exploit the semantic correlations between a question and its solution, we propose to maximize the mutual information between question-answer pairs and model-predicted solutions.
As demonstrated by \citet{DBLP:conf/emnlp/MinCHZ19}, for many QA tasks, it is feasible to precompute a modestly-sized, task-specific set of possible solutions containing the correct one. Therefore, we focus on handling the spurious solution problem under this circumstance. 
Specifically, we pair a task-specific model with a question reconstructor and repeat the following training cycle (Fig \ref{fig:train_cycle}): 
(1) sample a solution from the solution set according to model confidence, train the question reconstructor to reconstruct the question from that solution, and then (2) train the task-specific model on the most likely solution according to the question reconstructor. During training, the question reconstructor guides the task-specific model to predict those solutions consistent with the questions. For the question reconstructor, we devise an effective and unified way to encode solutions in different tasks, so that solutions with subtle differences (e.g., different spans with the same surface form) can be easily discriminated.

Our contributions are as follows: \textbf{(1)} We propose a mutual information maximization approach for the spurious solution problem in weakly supervised QA, which exploits the semantic correlations between a question and its solution; \textbf{(2)} We conducted extensive experiments on four QA datasets. Our approach significantly outperforms strong baselines in terms of task performance and is more effective in training models to produce correct solutions.
% achieves new state-of-the-art result on WikiSQL~\cite{DBLP:journals/corr/abs-1709-00103} under the weakly supervised setting.

\section{Related Work}
Question answering has raised prevalent attention and has achieved great progress these years. A lot of challenging datasets have been constructed to advance models' reasoning abilities, such as (1) reading comprehension datasets with extractive answer spans \cite{DBLP:conf/acl/JoshiCWZ17,DBLP:journals/corr/DhingraMC17}, with free-form answers \cite{DBLP:journals/tacl/KociskySBDHMG18}, for multi-hop reasoning \cite{DBLP:conf/emnlp/Yang0ZBCSM18}, or for discrete reasoning over paragraphs \cite{DBLP:conf/naacl/DuaWDSS019}, and (2) datasets for semantic parsing \cite{,DBLP:conf/acl/PasupatL15,DBLP:journals/corr/abs-1709-00103,DBLP:conf/emnlp/YuZYYWLMLYRZR18}. Under the weakly supervised setting, the specific solutions to derive the final answers (e.g., the correct location of an answer text, or the correct logic executing an answer) are not provided. This setting is worth exploration as it simplifies annotation and makes it easier to collect large-scale corpora. However, this setting introduces the spurious solution problem, and thus complicates model learning.

Most existing approaches for this learning challenge include heuristically selecting one possible solution per question for training \cite{DBLP:conf/acl/JoshiCWZ17,DBLP:conf/nips/TayLHS18,DBLP:conf/acl/TalmorB19}, training on all possible solutions with MML \cite{DBLP:conf/iclr/SwayamdiptaPK18,DBLP:conf/acl/GardnerC18,DBLP:conf/acl/LeeCT19,DBLP:conf/emnlp/WangTL19}, reinforcement learning \cite{DBLP:conf/acl/LiangBLFL17,DBLP:conf/nips/LiangNBLL18}, and hard EM \cite{DBLP:conf/emnlp/MinCHZ19,DBLP:conf/iclr/ChenLYZSL20}. All these approaches either use heuristics to select possibly reasonable solutions, rely on model architectures to bias towards correct solutions, or use model confidence to filter out spurious solutions in a soft or hard way. They do not explicitly exploit the semantic correlations between a question and its solution.

Most relevantly, \citet{DBLP:conf/conll/0001L18} focused on text2SQL tasks; they modeled SQL queries as the latent variables for question generation, and maximized the evidence lower bound of log likelihood of questions. A few works treated solution prediction and question generation as dual tasks and introduced dual learning losses to regularize learning under the fully-supervised or the semi-supervised setting \cite{DBLP:journals/corr/TangDQZ17,DBLP:conf/acl/CaoZLLY19,DBLP:conf/acl/YeLW19}. In dual learning, a model generates intermediate outputs (e.g., the task-specific model predicts solutions from a question) while the dual model gives feedback signals (e.g., the question reconstructor computes the likelihood of the question conditioned on predicted solutions). This method is featured in three aspects. First, both models need training on fully-annotated data so that they can produce reasonable intermediate outputs. Second, the intermediate outputs can introduce noise during learning as they are sampled from models but not restricted to solutions with correct answer or valid questions. Third, this method typically updates both models with reinforcement learning while the rewards provided by a dual model can be unstable or of high variance. By contrast, we focus on the spurious solution problem under the weakly supervised setting and propose a mutual information maximization approach. Solutions used for training are restricted to those with correct answer. What's more, though a task-specific model and a question reconstructor interact with each other, they do not use the likelihood from each other as rewards, which can stabilize learning.

\section{Method}

\subsection{Task Definition}
For a QA task, each instance is a tuple $\langle d$, $q$, $a\rangle$, where $q$ denotes a question, $a$ is the answer, and $d$ is reference information such as documents for reading comprehension, or table headers for semantic parsing. A solution $z$ is a task-specific derivation of the answer, e.g., a particular span in a document, an equation, or a SQL query (as shown in Fig \ref{fig:intro_case}). Let $f(\cdot)$ be the task-specific function that maps a solution to its execution result, e.g., by returning a particular span, solving an equation, or executing a SQL query. Our goal is to train a task-specific model $P_\theta(z|d, q)$ that takes $\langle d$, $q\rangle$ as input and predicts a solution $z$ satisfying $f(z)=a$.

Under the weakly supervised setting, only the answer $a$ is provided for training while the ground-truth solution $\bar{z}$ is not. We denote the set of possible solutions as $Z=\{z|f(z)=a\}$. In cases where the search space of solution is large, we can usually approximate $Z$ so that it contains the ground-truth solution $\bar{z}$ with a high probability \cite{DBLP:conf/emnlp/MinCHZ19,DBLP:conf/emnlp/WangTL19}. Note that $Z$ is task-specific, which will be instantiated in section \ref{sec:experiments}.

During training, we pair the task-specific model $P_\theta(z|d, q)$ with a question reconstructor $P_\phi(q|d, z)$ and maximize the mutual information between $\langle q$, $a \rangle$ and $z$.
During test, given $\langle d$, $q\rangle$, we use the task-specific model to predict a solution and return the execution result.

\begin{figure}[!t]
    \centering
    \includegraphics[width=0.48\textwidth]{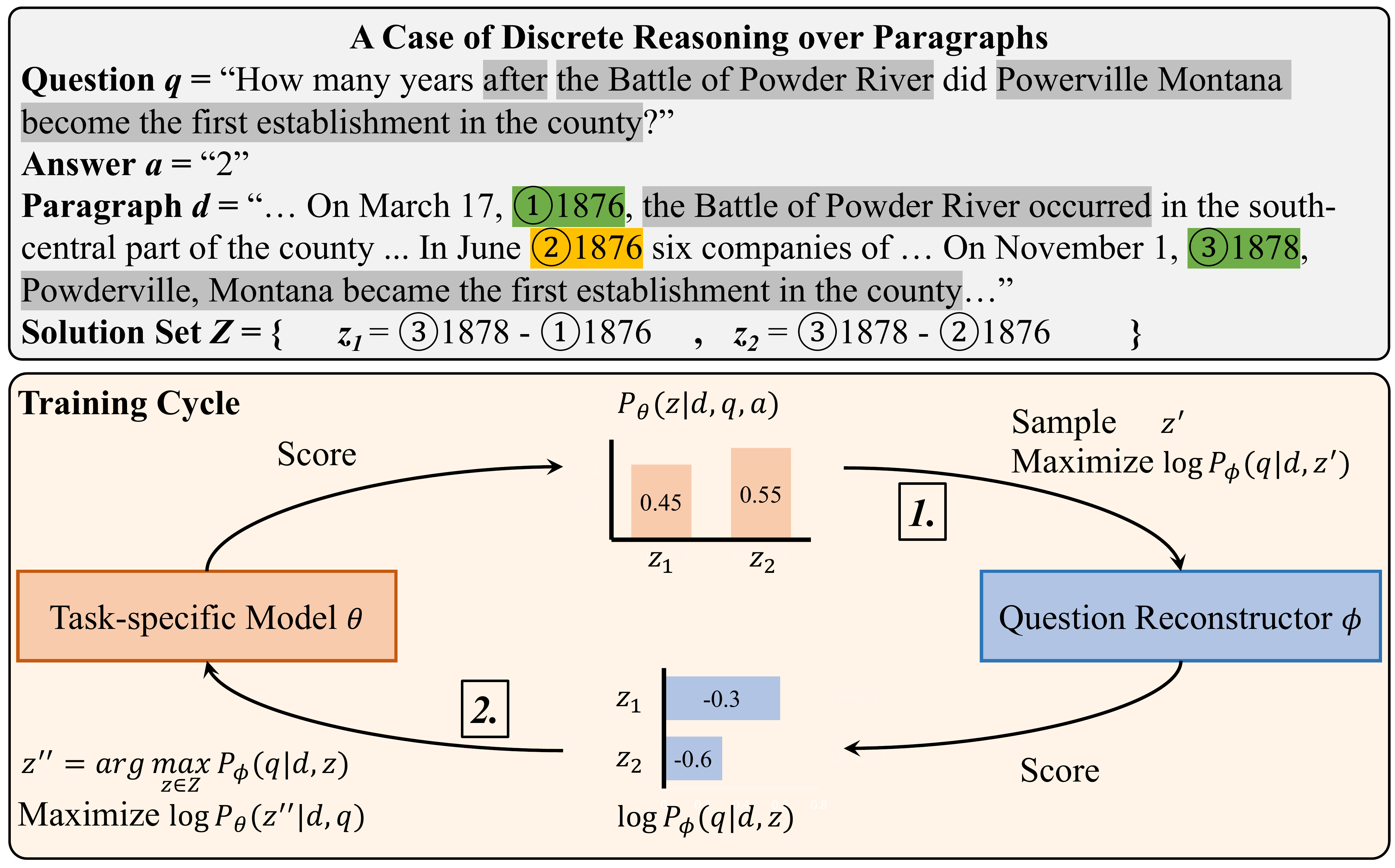}
    \caption{Illustration of the learning method.}
    \label{fig:train_cycle}
\end{figure}
\subsection{Learning Method}
%%%强烈建议有个总体框架图 ccc
Given an instance $\langle d$, $q$, $a\rangle$, the solution set $Z$ usually contains only one solution that best fits the instance while the rest are spurious. We propose to exploit the semantic correlations between a question and its solution to alleviate the spurious solution problem via mutual information maximization.

Our objective is to obtain the optimal task-specific model $\theta^*$ that maximizes the following conditional mutual information:
\begin{equation}
    \small
    \begin{split}
        \theta^* &= \arg\max_\theta I_\theta(\langle q, a \rangle; z|d) \\
        &= \arg\max_\theta H(\langle q, a\rangle|d) - H_\theta(\langle q, a\rangle|d, z) \\
        &= \arg\max_\theta -H_\theta(\langle q, a \rangle|d, z) \\
        & = \arg\max_\theta E_{P(d, q, a)}E_{P_\theta(z|d, q, a)} \log P_\theta(q, a|d, z)
    \end{split}
    \label{ori_obj}
\end{equation}
where $I_\theta(\langle q, a \rangle;z|d)$ denotes conditional mutual information between $\langle q, a \rangle$ and $z$ over $P(d, q, a)P_\theta(z|d, q, a)$. $H(\cdot|\cdot)$ is conditional entropy of random variable(s). $P(d, q, a)$ is the probability of an instance from the training distribution. $P_\theta(z|d, q, a)$ is the \textit{posterior prediction probability} of $z$ ($\in Z$) which is the prediction probability $P_\theta(z|d, q)$ normalized over $Z$:
\begin{equation}
    \small
    P_\theta(z|d, q, a) = 
    \begin{cases}
    \frac{P_\theta(z|d, q)}{\sum_{z^{'} \in Z}P_\theta(z^{'}|d, q)} & z \in Z \\
    0 & z \notin Z
    \end{cases}
    \label{post_pred}
\end{equation}

Note that computing $P_\theta(q, a|d, z)$ is intractable. We therefore introduce a question reconstructor $P_\phi(q|d, z)$ and approximate $P_\theta(q, a|d ,z)$ with $\mathbb{I}(f(z) = a)P_\phi(q|d, z)$ where $\mathbb{I}(\cdot)$ denotes indicator function. Eq. \ref{ori_obj} now becomes:
\begin{equation}
    \small
    \begin{split}
        \theta^* &= \arg\max_\theta \mathcal{L}_1 + \mathcal{L}_2 \\
        \mathcal{L}_1 &= E_{P(d, q, a)}E_{P_\theta(z|d, q, a)} \log P_\phi(q|d, z) \\
        \mathcal{L}_2 &= E_{P(d, q, a)}E_{P_\theta(z|d, q, a)} \log \frac{P_\theta(q, a|d, z)}{P_\phi(q|d, z)}
    \end{split}
    \label{new_obj}
\end{equation}
To optimize Eq. \ref{new_obj} is to repeat the following training cycle, which is analogous to the EM algorithm:
\begin{enumerate}
    % \item Compute the posterior prediction probability $P_\theta(z|d, q, a)$ for $z \in Z$ according to Eq. \ref{post_pred}.
    \item Minimize $\mathcal{L}_2$ w.r.t. the question reconstructor $\phi$ to draw $P_\phi(q|d, z)$ close to $P_\theta(q, a|d, z)$, by sampling a solution $z^{'} \in Z$ according to its posterior prediction probability $P_\theta(z|d, q, a)$ (see Eq. \ref{post_pred}) and maximizing $\log P_\phi(q|d, z^{'})$.
    \item Maximize $\mathcal{L}_1$ w.r.t. the task-specific model $\theta$. $\mathcal{L}_1$ can be seen as a reinforcement learning objective with $\log P_\phi(q|d, z)$ being the reward function. During training, the reward function is dynamically changing and may be of high variance. As we can compute the reward for all $z \in Z$, we therefore adopt a greedy but more stable update method, i.e., to maximize $\log P_\theta(z^{''}|d, q)$ where $z^{''} = \arg\max_{z \in Z} \log P_\phi(q|d, z)$ is the best solution according to the question reconstructor.
\end{enumerate}
We illustrate the above training cycle in Fig \ref{fig:train_cycle}.

\subsection{Question Reconstructor}
\begin{figure}[!t]
    \centering
    \includegraphics[width=0.48\textwidth]{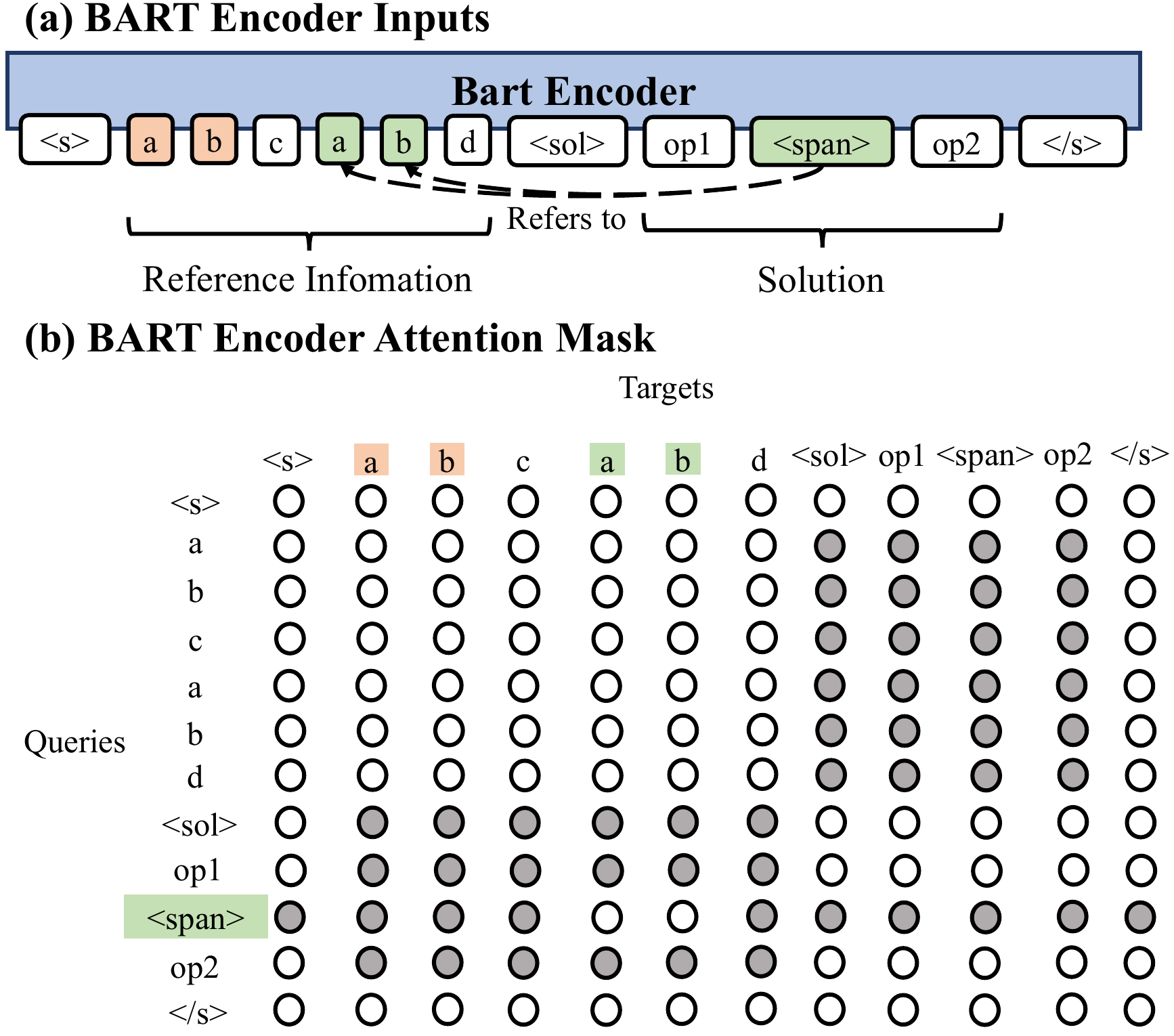}
    \caption{Solution encoding. (a) For BART encoder inputs, $\langle s \rangle$ and $\langle /s \rangle$ denote start and end of input sequence, respectively. $\langle sol \rangle$ denotes start of solution. $\langle span \rangle$ is the placeholder of the referred span in reference information (e.g., the second \textit{ab} in this figure. (b) For attention mask, gray circles block attention. $\langle span \rangle$ retrieves the contextual representation(s) of the referred span by only attending to the referred span. reference information and the solution (except for the token $\langle span \rangle$) are kept from attending to each other.}
    \label{fig:sol_enc}
\end{figure}
The question reconstructor $P_\phi(q|d, z)$ takes reference information $d$ and a solution $z$ as input, and reconstructs the question $q$. We use BART\textsubscript{base}, a pre-trained Seq2Seq model, as the question reconstructor so that semantic correlations between questions and solutions can be better captured. 

% However, it is problematic to just feed the concatenation of $d$ and the surface form of $z$ to the BART encoder. Take Figure **** as an example (**** positional information, contextual representation, spans with different meaning but the same surface form ****). To effectively encode $d$ and $z$, we devise a unified solution encoding as in Figure \ref{fig:sol_enc} which is applicable to solutions of different types in different tasks.

A solution typically consists of task-specific operation token(s) (e.g., COUNT for discrete reasoning or semantic parsing), literal(s) (e.g., numeric constants for discrete reasoning or semantic parsing), or span(s) from a question or reference information (e.g., for most QA tasks). It is problematic to just feed the concatenation of $d$ and the surface form of $z$ to the BART encoder; otherwise, different spans with the same surface form can no longer be discriminated as their contextual semantics are lost.
To effectively encode $d$ and $z$, we devise a unified solution encoding as in Fig \ref{fig:sol_enc} which is applicable to solutions of various types.
Specifically, we leave most of the surface form of $z$ unchanged, except that we replace any span from reference information with a placeholder $\langle span \rangle$. The representation of $\langle span \rangle$ is computed by forcing it to only attend to the contextual representation(s) of the referred span. To obtain disentangled and robust representations of reference information and a solution, we keep reference information and the solution (except for the token $\langle span \rangle$) from attending to each other. Intuitively speaking, semantics of reference information should not be affected by a solution, and the representations of a solution should largely determined by its internal logic.

\subsection{Solution Set}
While our learning method and question reconstructor are task-agnostic, solutions are usually task-specific. Precomputing solution sets needs formal definitions of solutions which define the search space of solutions. A possible search method is to exhaustively enumerate all solutions that produce the correct answer. 
We will introduce the definitions of solutions for different tasks in section \ref{sec:experiments}.

\section{Experiments}
\label{sec:experiments}
\begingroup
\setlength{\tabcolsep}{3pt} % Default value: 6pt
\renewcommand{\arraystretch}{1} % Default value: 1
\begin{table}[htb]
    \centering
	\adjustbox{max width=.45\textwidth}{
    \begin{tabular}{l|ccc|cc}
        \hline
        \multirow{2}{*}{Datasets} & \multicolumn{3}{c|}{\# Examples} & \multicolumn{2}{c}{$|Z|$} \\
        & Train & Dev & Test & Avg & Median \\
        \hline
        \multicolumn{6}{c}{Multi-mention Reading Comprehension} \\
        \hline
        Quasar-T & 37,012 & 3,000 & 3,000 & 8.1 & 4\\
        WebQuestions & 3,778 & - & 2,032 & 52.1 & 36\\
        \hline
        \multicolumn{6}{c}{Discrete Reasoning over Paragraphs} \\
        \hline
        DROP & 69,669 & 7,740 & 9,535 & 5.1 & 2\\
        \hline
        \multicolumn{6}{c}{Semantic Parsing} \\
        \hline
        WikiSQL & 56,355 & 8,421 & 15,878 & 315.4 & 4\\
	    \hline
    \end{tabular}
    }
    \caption{Statistics of the datasets we used. Statistics of the size of solution set $|Z|$ are computed on Train sets.}
    \label{tab:stat}
\end{table}
Following \citet{DBLP:conf/emnlp/MinCHZ19}, we conducted experiments on three QA tasks, namely multi-mention reading comprehension, discrete reasoning over paragraphs, and semantic parsing. This section introduces baselines, the definitions of solutions in different tasks, how the solution set can be precomputed, and our experimental results. Statistics of the datasets we used are presented in Table \ref{tab:stat}.

For convenience, we denote reference information as $d=[d_1, d_2, ..., d_{|d|}]$ and denote a question as $q=[q_1, q_2, ..., q_{|q|}]$ where $d_i$ and $q_j$ are a token of $d$ and $q$ respectively. A span from reference information and a question span is represented as $(s, e)^d$ and $(s, e)^q$ respectively, where $s$ and $e$ are start and end index of the span respectively.

\subsection{Baselines}
% \paragraph{First Only}
\textbf{First Only} \cite{DBLP:conf/acl/JoshiCWZ17}
which trains a reading comprehension model by maximizing $\log P_\theta(z|d, q)$ where $z$ is the first answer span in $d$.

% \paragraph{MML}
\noindent
\textbf{MML} \cite{DBLP:conf/emnlp/MinCHZ19}
which maximizes $\log \sum_{z \in Z} P_\theta(z|d, q)$.

% \paragraph{HardEM}
\noindent
\textbf{HardEM} \cite{DBLP:conf/emnlp/MinCHZ19}
which maximizes $\log max_{z \in Z} P_\theta(z|d, q)$.

% \paragraph{HardEM-thres}
\noindent
\textbf{HardEM-thres} \cite{DBLP:conf/iclr/ChenLYZSL20}:
a variant of HardEM that optimizes only on confident solutions, i.e., to maximize $max_{z \in Z} \mathbb{I}(P_\theta(z|d, q) > \gamma)\log P_\theta(z|d, q)$ where $\gamma$ is an exponentially decaying threshold. $\gamma$ is initialized such that a model is trained on no less than half of training data at the first epoch. We halve $\gamma$ after each epoch.

% \paragraph{VAE}
\noindent
\textbf{VAE} \cite{DBLP:conf/conll/0001L18}:
a method that views a solution as the latent variable for question generation and adopts the training objective of Variational Auto-Encoder (VAE) \cite{DBLP:journals/corr/KingmaW13} to regularize the task-specific model. The overall training objective is given by:
\begin{equation}
    \small
    \notag
    \begin{split}
        \theta^*, \phi^* &= \arg\max_{\theta, \phi} \mathcal{L}(\theta, \phi) \\
        \mathcal{L}(\theta, \phi) &= \mathcal{L}^{mle}(\theta) + \lambda \mathcal{L}^{vae}(\theta, \phi) \\
        &=\sum_{z \in B} \log P_\theta(z|d, q) + \lambda E_{P_\theta(z|d, q)} \log \frac{P_\phi(q|d, z)}{P_\theta(z|d, q)} \\
    \end{split}
\end{equation}
where $\theta$ denotes a task-specific model and $\phi$ is our question reconstructor. $\mathcal{L}^{mle}(\theta)$ is the total log likelihood of the set of model-predicted solutions (denoted by $B$) which derive the correct answer. $\mathcal{L}^{vae}(\theta, \phi)$ is the evidence lower bound of the log likelihood of questions. $\lambda$ is the coefficient of $\mathcal{L}^{vae}(\theta, \phi)$. This method needs pre-training both $\theta$ and $\phi$ before optimizing the overall objective $\mathcal{L}(\theta, \phi)$. Notably, model $\theta$ optimizes on $\mathcal{L}^{vae}(\theta, \phi)$ via reinforcement learning. We tried stabilizing training by reducing the variance of rewards and setting a small $\lambda$.

\subsection{Multi-Mention Reading Comprehension}
Multi-mention reading comprehension is a natural feature of many QA tasks. 
Given a document $d$ and a question $q$, a task-specific model is required to locate the answer text $a$ which is usually mentioned many times in the document(s). A solution is defined as a document span. The solution set $Z$ is computed by finding exact match of $a$:
\begin{equation}
    \small
    \notag
    \begin{split}
        % g_{max} &= \max_{1\le s \le e \le |d|} g([d_s, ..., d_e], a)\\
        Z &= \{z=(s, e)^d|[d_s, ..., d_e] = a\}
    \end{split}
\end{equation}
% where $g(\cdot, \cdot)$ is Exact Match (EM) for extractive QA (e.g., Quasar-T \cite{DBLP:journals/corr/DhingraMC17} where $a$ is a document span) and ROUGE-L \cite{lin-2004-rouge} for abstractive QA (e.g., NarrativeQA \cite{DBLP:journals/tacl/KociskySBDHMG18} where $a$ is a free-form text).

We experimented on two open domain QA datasets, i.e., Quasar-T \cite{DBLP:journals/corr/DhingraMC17} and WebQuestions \cite{DBLP:conf/emnlp/BerantCFL13}. 
For Quasar-T, we retrieved 50 reference sentences from ClueWeb09 for each question; for WebQuestions, we used the 2016-12-21 dump of Wikipedia as the knowledge source and retrieved 50 reference paragraphs for each question using a Lucene index system.
We used the same BERT\textsubscript{base} \cite{DBLP:conf/naacl/DevlinCLT19} reading comprehension model and data preprocessing from \cite{DBLP:conf/emnlp/MinCHZ19}.
\begingroup
\setlength{\tabcolsep}{3pt} % Default value: 6pt
\renewcommand{\arraystretch}{1} % Default value: 1
\begin{table}[htb]
    \centering
	\adjustbox{max width=.45\textwidth}{
    \begin{tabular}{c|cc|cc|cc}
        \hline
		& \multicolumn{4}{c|}{Quasar-T} & \multicolumn{2}{c}{WebQuestions} \\
		\cline{2-7}
		& \multicolumn{2}{c|}{Dev} & \multicolumn{2}{c|}{Test} & \multicolumn{2}{c}{Test} \\
		& EM & F1 & EM & F1 & EM & F1 \\
        \hline
        First Only & 36.0 & 43.9 & 35.6 & 42.8 & 16.7 & 22.6 \\
        % \hline
        MML & 40.1 & 47.4 & 39.1 & 46.5 & 18.4 & 25.0 \\
        % \hline
        HardEM & 41.5 & 49.1 & 40.7 & 47.7 & 18.0 & 24.2 \\
        % \hline
        HardEM-thres & 42.8 & 50.2 & 41.9 & 49.4 & 19.0 & 25.3 \\
        \hline
        Ours & \textbf{44.7}\textsuperscript{$\ddag$} & \textbf{52.6}\textsuperscript{$\ddag$} & \textbf{44.0}\textsuperscript{$\ddag$} & \textbf{51.5}\textsuperscript{$\ddag$} & \textbf{20.4}\textsuperscript{$\ddag$} & \textbf{27.2}\textsuperscript{$\ddag$} \\
	    \hline
    \end{tabular}
    }
    \caption{Evaluation on multi-mention reading comprehension datasets. Numbers marked with \textsuperscript{$\ddag$} are significantly better than the others (t-test, p-value $<$ 0.05).}
    \label{tab:rc_res}
\end{table}
\begingroup
\setlength{\tabcolsep}{3pt} % Default value: 6pt
\renewcommand{\arraystretch}{1} % Default value: 1
\begin{table*}[!tp]
    \centering
	\adjustbox{max width=0.75\textwidth}{
    \begin{tabular}{c|cc|cc|cc|cc|cc}
        \hline
		& \multicolumn{2}{c|}{Overall Test} & \multicolumn{2}{c|}{Number (61.97\%)} & \multicolumn{2}{c|}{Span (31.47\%)} & \multicolumn{2}{c|}{Spans (4.99\%)} & \multicolumn{2}{c}{Date (1.57\%)} \\
		& EM & F1 & EM & F1 & EM & F1 & EM & F1 & EM & F1 \\
		\hline
		MML & 58.99\textsuperscript{$\ddag$} & 62.30\textsuperscript{$\ddag$} & 55.38 & 55.58 & 69.96 & 75.51 & 39.29 & 66.01 & 42.57 & 49.05\\
		HardEM & 68.52\textsuperscript{$\ddag$} & 71.88\textsuperscript{$\ddag$} & 68.40 & 68.70 & 73.50 & 79.25 & 44.79 & 69.63 & 49.32 & 56.87 \\
		HardEM-thres & 69.06 & 72.35\textsuperscript{$\ddag$} & 69.05 & 69.39 & 74.61 & 79.79 & 39.50 & 66.38 & 52.67 & 58.75\\
		VAE & 32.34\textsuperscript{$\ddag$} & 36.28\textsuperscript{$\ddag$} & 51.65 & 52.35 & 0.37 & 10.01 & 0.00 & 8.89 & 0.00 & 4.11\\
	    \hline
	    Ours & \textbf{69.35} & \textbf{72.92} & \textbf{69.96} & \textbf{70.27} & 73.38 & 79.32 & 42.86 & \textbf{70.42} & 48.67 & 57.47\\
	    \hline
    \end{tabular}
    }
    \caption{Evaluation on DROP. We used the public development set of DROP as our test set. We also provide performance breakdown of different question types on our test set. Results on the overall test set marked with \textsuperscript{$\ddag$} are significantly worse than the best one (t-test, p-value $<$ 0.05).}
    \label{tab:drop_res}
\end{table*}

% \paragraph{Results}
\noindent
\textbf{Results:} Our method outperforms all baselines on both datasets (Table \ref{tab:rc_res}). The improvements can be attributed to the effectiveness of solution encoding, as solutions for this task are typically different spans with the same surface form, e.g., in Qusart-T, all $z \in Z$ share the same surface form.

\subsection{Discrete Reasoning over Paragraphs}
\begin{table}[htb]
\centering
\renewcommand{\arraystretch}{1.3}
\adjustbox{max width=.45\textwidth}{
\begin{tabular}{c|c}
\hline
\multicolumn{2}{c}{Numeric Answers} \\
\hline
Arithmetic & {\small $\begin{aligned} z=&n_1 [, o_1, n_2 [, o_2, n_3]], \\ \text{s.t. }&o_1, o_2 \in \{+, -\},\\ &n_1, n_2, n_3 \in N_d \cup S \end{aligned}$}\\
\hline
Sorting & {\small $\begin{aligned}z = &o\{n_k\}_{k \ge 1},\\ \text{s.t. }&o \in \{max, min\}, n_k \in N_d \end{aligned}$} \\
\hline
Counting & {\small $\begin{aligned} z = &|\{(s_k, e_k)^d\}_{k \ge 1}|\end{aligned}$} \\
\hline
\multicolumn{2}{c}{Non-numeric Answers} \\
\hline
Span(s) & {\small $\begin{aligned} z = \{(s_k, e_k)^t\}_{k \ge 1}, \text{s.t. } t \in \{d, q\}\end{aligned}$} \\
\hline
Sorting & {\small $\begin{aligned}z = &o\{kv\langle(s_k, e_k)^d, n_k\rangle\}_{k \ge 1},\\ \text{s.t. }&o \in \{argmax, argmin\}, n_k \in N_d \end{aligned}$} \\
\hline
\end{tabular}}
\caption{Definitions of solutions for numeric answers and non-numeric answers. $N_d$ is the set of numbers in $d$, and $S$ is a set of pre-defined numbers. For arithmetic solutions for numeric answers, $z = n_1[,o_1, n_2[,o_2, n_3]]$ denotes equations with no more than three operands. For solutions of sorting type for non-numeric answers, $kv\langle\cdot, \cdot\rangle$ is a key-value pair where the 
key is a span in $d$ and the value is its associated number from $d$. $argmax$ ($argmin$) returns the key with the largest (smallest) value.}
\label{tab:drop_sol}
\end{table}
Some reading comprehension tasks pose the challenge of comprehensive analysis of texts by requiring discrete reasoning (e.g., arithmetic calculation, sorting, and counting) \cite{DBLP:conf/naacl/DuaWDSS019}.
In this task, given a paragraph $d$ and a question $q$, an answer $a$ can be one of the four types: numeric value, a paragraph span or a question span, a sequence of paragraph spans, and a date from the paragraph. The definitions of $z$ depend on answer types (Table \ref{tab:drop_sol}). 
These solutions can be searched by following \citet{DBLP:conf/iclr/ChenLYZSL20}. 
Note that some solutions involve numbers in $d$. We treated those numbers as spans while reconstructing $q$ from $z$.

We experimented on DROP \cite{DBLP:conf/naacl/DuaWDSS019}. As the original test set is hidden, for convenience of analysis, we used the public development set as our test set, and split the public train set into 90\%/10\% for training and development. We used Neural Symbolic Reader (NeRd) \cite{DBLP:conf/iclr/ChenLYZSL20} as the task-specific model. NeRd is a Seq2Seq model which encodes a question and a paragraph, and decodes a solution (e.g., \textit{count (paragraph\_span($s_1$, $e_1$), paragraph\_span($s_2$, $e_2$))} where \textit{paragraph\_span($s_i$, $e_i$)} means a paragraph span starting at $s_i$ and ending at $e_i$). We used the precomputed solution sets provided by \citet{DBLP:conf/iclr/ChenLYZSL20}\footnote{Our implementation of NeRd has four major differences 
from that of \cite{DBLP:conf/iclr/ChenLYZSL20}. (1) Instead of choosing BERT\textsubscript{large} as encoder, we chose the discriminator of Electra\textsubscript{base} \cite{DBLP:conf/iclr/ClarkLLM20} which is of a smaller size. (2) We did not use moving averages of trained parameters. (3) We did not use the full public train set for training but used 10\% of it for development. (4) For some questions, it is hard to guarantee that a precomputed solution set covers the ground-truth solution. For example, the question \textit{How many touchdowns did Brady throw?} needs counting, but the related mentions are not known. \cite{DBLP:conf/iclr/ChenLYZSL20} partly solved this problem by adding model-predicted solutions (with correct answer) into the initial solution sets as learning proceeds. In this paper, we kept the initial solution sets unchanged during training, so that different QA tasks share the same experimental setting.}. 
Data preprocessing was also kept the same.

% \paragraph{Results}
\noindent
\textbf{Results:}
As shown in Table \ref{tab:drop_res}, our method significantly outperforms all baselines in terms of F1 score on our test set.

We also compared our method with the baseline VAE which uses a question reconstructor $\phi$ to adjust the task-specific model $\theta$ via maximizing a variational lower bound of $\log P(q|d)$ as the regularization term $\mathcal{L}^{vae}(\theta, \phi)$. 
To pre-train the task-specific model for this method, we simply obtained the best task-specific model trained with HardEM-thres. VAE optimizes the task-specific model on $\mathcal{L}^{vae}(\theta, \phi)$ with reinforcement learning where $P_\phi(q|d, z)$ is used as learning signals for the task-specific model. Despite our efforts to stabilize training, the F1 score still dropped to 36.28 after optimizing the overall objective $\mathcal{L}(\theta, \phi)$ for 1,000 steps. By contrast, our method does not use $P_\phi(q|d, z)$ to compute learning signals for the task-specific model but rather uses it to select solutions to train the task-specific model, which makes a better use of the question reconstructor.

\subsection{Semantic Parsing}
Text2SQL is a popular semantic parsing task. Given a question $q$ and a table header $d=[h_1, ..., h_L]$ where $h_l$ is a multi-token column, a parser is required to parse $q$ into a SQL query $z$ and return the execution results. Under the weakly supervised setting, only the final answer is provided while the SQL query is not.
Following \citet{DBLP:conf/emnlp/MinCHZ19}, $Z$ is approximated as a set of non-nested SQL queries with no more than three conditions:
\begin{equation}
    \small
    \notag
    \begin{split}
        Z = \{z &= (z^{sel}, z^{agg}, \{z_k^{cond}\}_{k=1}^3) | f(z) = a, \\
        z^{sel} &\in \{h_1, ..., h_L\}, z_k^{cond} \in \{none\} \cup C,\\
        z^{agg} &\in \{none, sum, mean, max, min, count\} \}
    \end{split}
\end{equation}
where $z^{agg}$ is an aggregating operator and $z^{sel}$ is the operated column (a span of $d$). $C=\{(h, o, v)\}$ is the set of all possible conditions, where $h$ is a column, $o \in \{=, <, >\}$, and $v$ is a question span.

We experimented on WikiSQL \cite{DBLP:journals/corr/abs-1709-00103} under the weakly supervised setting\footnote{WikiSQL has annotated ground-truth SQL queries. We only used them for evaluation but not for training.}. We chose SQLova \cite{DBLP:journals/corr/abs-1902-01069} as the task-specific model which is a competitive text2SQL parser on WikiSQL. Hyperparameters were kept the same as in \cite{DBLP:journals/corr/abs-1902-01069}. We used the solution sets provided by \citet{DBLP:conf/emnlp/MinCHZ19}.

% \paragraph{Results}
\noindent
\textbf{Results}: All models in Table \ref{tab:wikisql_res} do not apply execution-guided decoding during inference. Our method achieves new state-of-the-art results under the weakly supervised setting. Though without supervision of ground-truth solutions, our execution accuracy (i.e., accuracy of execution results) on the test set is close to that of the fully supervised SQLova. Notably, GRAPPA focused on representation learning and used a stronger task-specific model while we focus on the learning method and outperform GRAPPA with a weaker model.

\section{Ablation Study}
\subsection{Performance on Test Data with Different Size of Solution Set}
Fig \ref{fig:bd} shows the performance on test data with different size of solution set\footnote{In this experiment, $|Z|$ is only seen as a property of an example. Evaluated solutions are predicted by the task-specific model but not from $Z$.}. Our method consistently outperforms HardEM-thres and by a large margin when test examples have a large solution set.
% \begin{figure}[!tp]
%      \centering
%      \begin{subfigure}[b]{0.49\textwidth}
%          \centering
%          \includegraphics[width=\textwidth]{DROP_breakdown.pdf}
%          \caption{}
%          \label{fig:drop_bd}
%      \end{subfigure}
%      \hfill
%      \begin{subfigure}[b]{0.49\textwidth}
%          \centering
%          \includegraphics[width=\textwidth]{WikiSQL_breakdown.pdf}
%          \caption{}
%          \label{fig:sql_bd}
%      \end{subfigure}
%     \caption{Performance breakdown of test examples with different size of $Z$ on (a) DROP and (b) WikiSQL.}
%     \label{fig:bd}
% \end{figure}

\subsection{Effect of $|Z|$ at Training}
The more complex a question is, the larger the set of possible solutions tends to be, the more likely a model will suffer from the spurious solution problem. We therefore investigated whether our learning method can deal with extremely noisy solution sets. Specifically, we extracted a hard train set from the original train set of WikiSQL. The hard train set consists of 10K training data with the largest $Z$. The average size of $Z$ on the hard train set is 1,554.6, much larger than that of the original train set (315.4). We then compared models trained on the original train set and the hard train set using different learning methods.

As shown in Fig \ref{fig:sql}, models trained with our method consistently outperform baselines in terms of logical form accuracy (i.e., accuracy of predicted solutions) and execution accuracy. When using the hard train set, the logical form accuracy of models trained with HardEM or HardEM-thres drop to below 14\%. Compared with HardEM, HardEM-thres is better when trained on the original train set but is worse when trained on the hard train set. These indicate that model confidence can be unreliable and thus insufficient to filter out spurious solutions. By contrast, our method explicitly exploits the semantic correlations between a question and a solution, thus much more resistant to spurious solutions.
\begingroup
\setlength{\tabcolsep}{3pt} % Default value: 6pt
\renewcommand{\arraystretch}{1} % Default value: 1
\begin{table}[!tp]
    \centering
	\adjustbox{max width=.4\textwidth}{
    \begin{tabular}{lcc}
        \hline
        \multirow{2}{*}{Model} & \multicolumn{2}{c}{Execution Accuracy} \\
        & Dev & Test \\
        \hline
        \multicolumn{3}{c}{\textbf{Fully-supervised Setting}} \\
	    \hline
	   % Coarse2Fine \cite{DBLP:conf/acl/LapataD18} & 79.0 & 78.5 \\
	    SQLova \cite{DBLP:journals/corr/abs-1902-01069} & 87.2 & 86.2 \\
	   % X-SQL \cite{DBLP:journals/corr/abs-1908-08113} & 89.5 & 88.7\\
	   % IE-SQL \cite{ma-etal-2020-mention} & 88.7 & 88.8\\
	    HydraNet \cite{DBLP:journals/corr/abs-2008-04759} & \textbf{89.1} & \textbf{89.2}\\
	    \hline
	    \multicolumn{3}{c}{\textbf{Weakly-supervised Setting}} \\
	    \hline
	   % MAPO \cite{DBLP:conf/nips/LiangNBLL18} & 71.8 & 72.4 \\
	    MeRL \cite{DBLP:conf/icml/AgarwalLS019} & 74.9 & 74.8 \\
	    GRAPPA \cite{yu2021grappa} & 85.9 & 84.7 \\
	    \hline
	    MML\cite{DBLP:conf/emnlp/MinCHZ19} & 70.6 & 70.5 \\
	    HardEM & 84.5\textsuperscript{$\ddag$} & 84.1\textsuperscript{$\ddag$} \\
	    HardEM-thres & 85.2\textsuperscript{$\dag$} & 84.1\textsuperscript{$\ddag$} \\
	    Ours & \textbf{85.9} & \textbf{85.6}\\
	    \hline
    \end{tabular}
    }
    \caption{Evaluation on WikiSQL. Accuracy that is significantly lower than the highest one is marked with \textsuperscript{$\dag$} for p-value $<$ 0.1, and \textsuperscript{$\ddag$} for p-value $<$ 0.05 (t-test).}
    \label{tab:wikisql_res}
\end{table}
\begin{figure}[!t]
    \centering
    \includegraphics[width=0.45\textwidth]{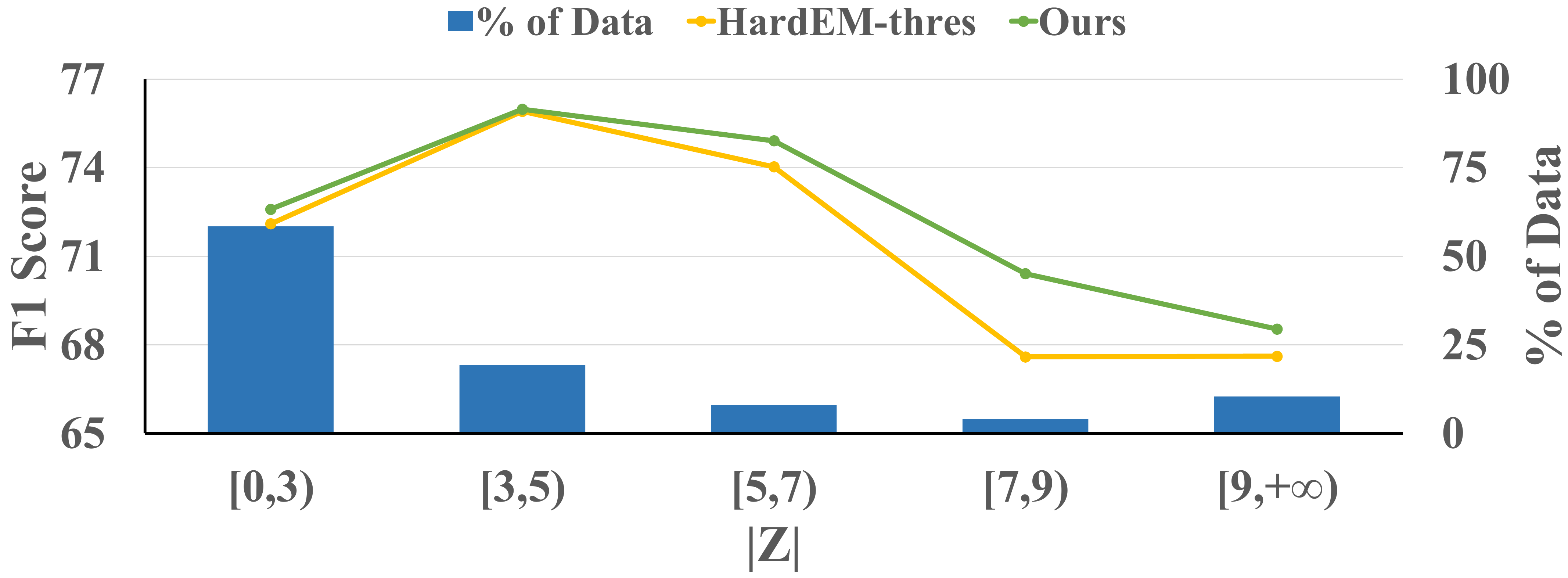}
    \caption{Performance on test examples with different size of $Z$ on DROP.}
    \label{fig:bd}
\end{figure}
\begin{figure}[!t]
    \centering
    \includegraphics[width=0.48\textwidth]{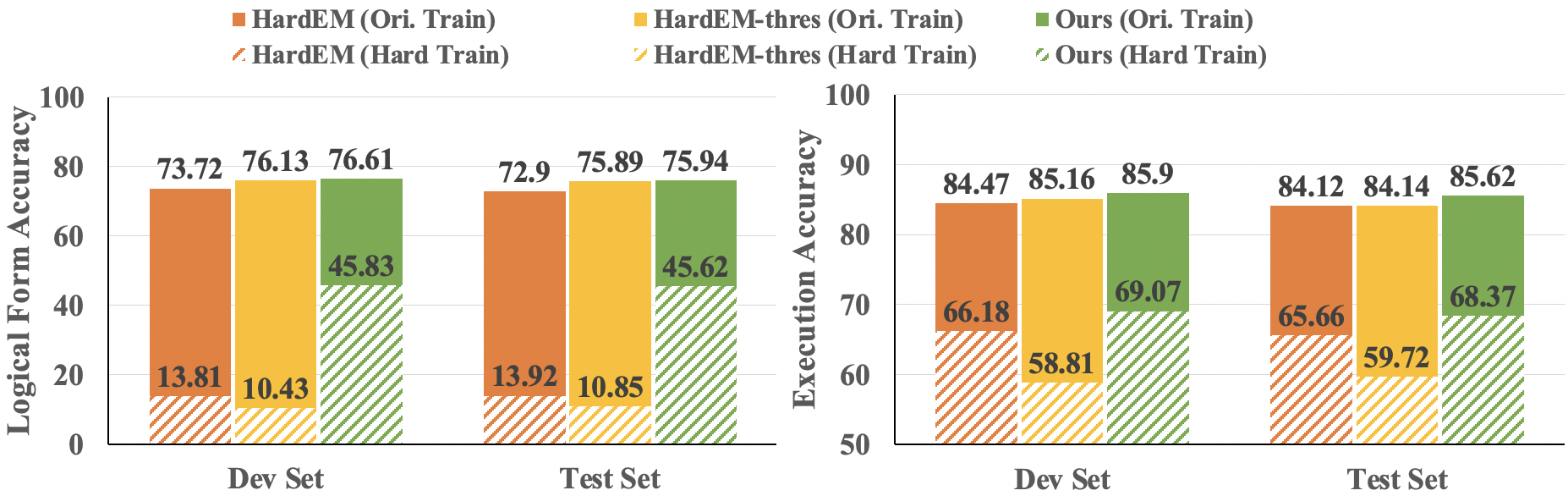}
    \caption{Logical form accuracy (left) and execution accuracy (right) on dev set and test set of WikiSQL. A method marked with \textit{Ori. Train} or \textit{Hard Train} means the evaluated model is trained on the original train set or a hard subset of training data, respectively. The hard train set consists of 10K training data with the largest solution set; the average size of solution set is 1,554.6.}
    \label{fig:sql}
\end{figure}

\subsection{Effect of the Question Reconstructor}
As we used BART\textsubscript{base} as the question resconstructor, we investigated how our question reconstructor contributes to performance improvements.

We first investigated whether BART\textsubscript{base} itself is less affected by the spurious solution problem than the task-specific models. Specifically, we viewed text2SQL as a sequence generation task and finetuned a BART\textsubscript{base} on the hard train set of WikiSQL with HardEM. The input of BART shares the same format as that of SQLova, which is the concatenation of a question and a table header. The output of BART is a SQL query. Without constraints on decoding, BART might not produce valid SQL queries. We therefore evaluated models on a SQL selection task instead: for each question in the development set of WikiSQL, a model picks out the correct SQL from at most 10 candidates by selecting the one with the highest prediction probability. As shown in Table \ref{tab:sql_selection}, when trained with HardEM, both BART\textsubscript{base} parser and SQLova perform similarly, and underperform our method by a large margin. This indicates that using BART\textsubscript{base} as a task-specific model can not avoid the spurious solution problem. It is our mutual information maximization objective that makes a difference.
\begingroup
\setlength{\tabcolsep}{3pt} % Default value: 6pt
\renewcommand{\arraystretch}{1} % Default value: 1
\begin{table}[!tp]
    \centering
	\adjustbox{max width=.35\textwidth}{
    \begin{tabular}{c|c|c|c|c|c}
        \hline
        Training Epochs & 2 & 4 & 6 & 8 & 10 \\
        \hline
        BART\textsubscript{base} w/ HardEM & 65.1 & 60.8 & 59.7 & 58.6 & 61.0\\
        SQLova w/ HardEM & 61.3 & 62.2 & 61.8 & 61.8 & 61.7\\
        \hline
        SQLova w/ Ours & \textbf{79.7} & \textbf{82.8} & \textbf{79.8} & \textbf{81.2} & \textbf{87.4}\\
	    \hline
    \end{tabular}
    }
    \caption{Accuracy on the SQL selection task. The hard train set was used for training. \textit{BART\textsubscript{base} w/ HardEM} and \textit{SQLova w/ HardEM} are a BART\textsubscript{base} parser and SQLova, respectively; both were trained with HardEM. \textit{SQLova w/ Ours} is SQLova trained with the proposed mutual information maximization approach (using BART\textsubscript{base} question reconstructor).}
    \label{tab:sql_selection}
\end{table}

\begingroup
\setlength{\tabcolsep}{3pt} % Default value: 6pt
\renewcommand{\arraystretch}{1} % Default value: 1
\begin{table}[htp]
    \centering
	\adjustbox{max width=.45\textwidth}{
    \begin{tabular}{c|cc|cc|cc|cc}
        \hline
        & \multicolumn{4}{c|}{DROP} & \multicolumn{4}{c}{WikiSQL (Hard Train Set)} \\
        \cline{2-9}
        & \multicolumn{2}{c|}{Dev} & \multicolumn{2}{c|}{Test} & \multicolumn{2}{c|}{Dev} & \multicolumn{2}{c}{Test} \\
        & EM & F1 & EM & F1 & LF. Acc & Exe. Acc & LF. Acc & Exe. Acc \\
        \hline
        T-scratch & \textbf{61.5} & 66.3 & 69.0 & 72.4 & 24.7 & 67.9 & 24.9 & 67.5\\
        T-DAE & \textbf{61.5} & 66.3 & \textbf{69.4} & 72.7 & \textbf{49.4} & 68.9 & \textbf{48.5} & \textbf{68.4}\\
        BART\textsubscript{base} & \textbf{61.5} & \textbf{66.4} & 69.3 & \textbf{72.9} & 45.8 & \textbf{69.1} & 45.6 & \textbf{68.4} \\
	    \hline
    \end{tabular}
    }
    \caption{Results with different question reconstructors. \textit{LF. Acc} and \textit{Exe. Acc} are logical form accuracy and execution accuracy, respectively. \textit{T-scratch} is a Transformer without pre-training. \textit{T-DAE} is a Transformer pre-trained as a denoising auto-encoder of questions.}
    \label{tab:sql_que_recon}
\end{table}

We further investigated the effect of the choice of question reconstructor. We compared BART\textsubscript{base} with two alternatives: (1) \textbf{T-scratch:} a three-layer Transformer \cite{DBLP:conf/nips/VaswaniSPUJGKP17} without pre-training and (2) \textbf{T-DAE:} a three-layer Transformer pre-trained as a denoising auto-encoder of questions on the train set; the text infilling pre-training task for BART was used. As shown in Table \ref{tab:sql_que_recon}, our method with either of the three question reconstructors outperforms or is at least competitive with baselines, which verifies the effectiveness of our mutual information maximization objective. What's more, using T-DAE is competitive with BART\textsubscript{base}, indicating that our training objective is compatible with other choices of question reconstructor besides BART, and that using a denoising auto-encoder to initialize the question reconstructor may be beneficial to exploit the semantic correlations between a question and its solution.

\section{Evaluation of Solution Prediction}
As solutions with correct answer can be spurious, we further analyzed the quality of predicted solutions. We randomly sampled 50 test examples from DROP for which our method produced the correct answer, and found that our method also produced the correct solution for 92\% of them.

To investigate the effect of different learning methods on models' ability to produce correct solutions, we manually analyzed another 50 test samples for which HardEM, HardEM-thres, and our method produced the correct answer with different solutions. The percentage of samples for which our method produced the correct solution is 58\%, much higher than that of HardEM (10\%) and HardEM-thres (30\%).
\textbf{For experimental details, please refer to the appendix.}

\section{Case Study}
\begin{figure*}[!t]
    \centering
    \includegraphics[width=0.95\textwidth]{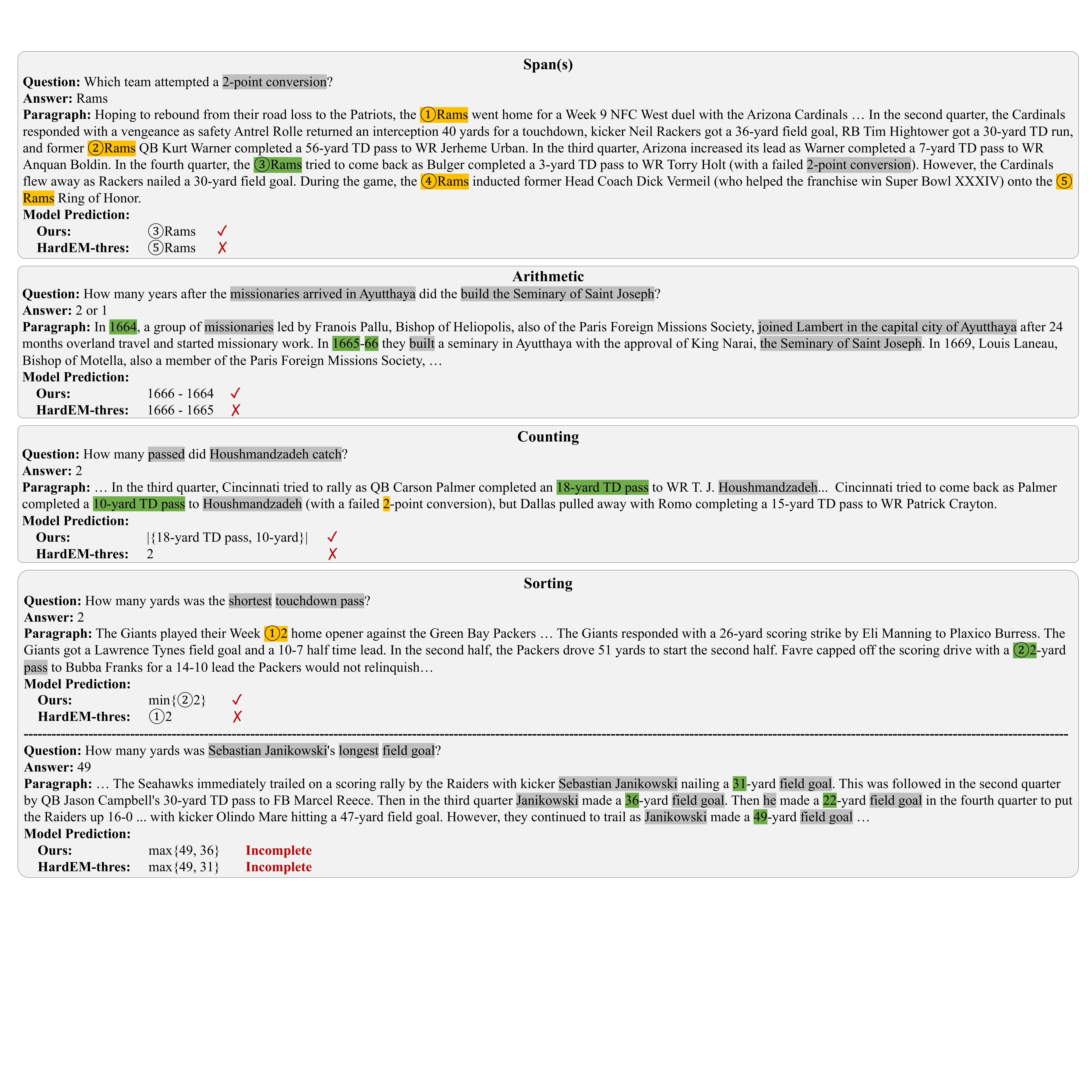}
    \caption{NeRd predictions on four types of questions from DROP when using different learning methods. Spans in dark gray and green denote semantic correlations between a question and its solution, while spans in orange are spurious information and should not be used in a solution.}
    \label{fig:drop_cases}
\end{figure*}
Fig \ref{fig:drop_cases} compares NeRd predictions on four types of questions from DROP when using different learning methods. An observation is that NeRd using our method shows more comprehensive understanding of questions, e.g., in the \textit{Arithmetic} case, NeRd using our method is aware of the two key elements in the question including the year when \textit{missionaries arrived in Ayutthaya} and the year when \textit{the Seminary of Saint Joseph} was built, while NeRd using HardEM-thres misses the first element. What's more, NeRd using our method is more precise in locating relevant information, e.g., in the first \textit{Sorting} case, NeRd with our method locates the second appearance of \textit{2} whose contextual semantics matches the question, while NeRd using HardEM-thres locates the first appearance of \textit{2} which is irrelevant. These two observations can be attributed to our mutual information maximization objective which biases a task-specific model towards those solutions that align well with the questions.

However, we also observed that when there are multiple mentions of relevant information of the same type, NeRd trained with HardEM-thres or our method has difficulty in recalling them all, e.g., in the second \textit{Sorting} case, the correct solution should locate all four mentions of \textit{Sebastian Janikowski's field goals} while NeRd using either method locates only two of them. We conjecture that this is because the solution sets provided by \citet{DBLP:conf/iclr/ChenLYZSL20} are noisy. For example, all precomputed solutions of sorting type for numeric answers involve up to two numbers from reference information, which makes it hard for a model to learn to sort more than two numbers.

\section{Conclusion}
To alleviate the spurious solution problem in weakly supervised QA, we propose to explicitly exploit the semantic correlations between a question and its solution via mutual information maximization. During training, we pair a task-specific model with a question reconstructor which guides the task-specific model to predict solutions that are consistent with the questions. Experiments on four QA datasets demonstrate the effectiveness of our learning method.
As shown by automatic and manual analyses, models trained with our method are more resistant to spurious solutions during training, and are more precise in locating information that is relevant to the questions during inference, leading to higher accuracy of both answers and solutions.

\section{Acknowledgements}
This work was partly supported by the NSFC projects (Key project with No. 61936010 and regular project with No. 61876096). This work was also supported by the Guoqiang Institute of Tsinghua University, with Grant No. 2019GQG1 and 2020GQG0005.

\newpage

\bibliographystyle{acl_natbib}
\bibliography{acl2021-camera-ready}

\newpage
\appendix

\section{Implementation Details}
\subsection{Learning Methods}
\noindent
\textbf{HardEM}: We followed \citet{DBLP:conf/emnlp/MinCHZ19} to apply annealing to HardEM on reading comprehension tasks: at the training step $t$, a model optimizes MML objective with a probability of $\min(t / \tau, 0.8)$ and optimizes HardEM objective otherwise. $\tau$ was chosen from $\{10K, 20K, 30K, 40K, 50K\}$ based on model performance on the development set.

\noindent
\textbf{HardEM-thres}: We set the confidence threshold as $\gamma = 0.5^{n}$ where $n$ was initialized as follows: we first computed the prediction probability of each solution with a task-specific model, and then set $n$ to a value such that the model was trained on no less than half of training data at the first epoch. We halved $\gamma$ after each epoch.

\noindent
\textbf{VAE}\cite{DBLP:conf/conll/0001L18}: A method that views a solution as the latent variable for question generation and adopts the training objective of Variational Auto-Encoder (VAE) to regularize the task-specific model. The overall training objective is given by:
\begin{equation}
    \small
    \notag
    \begin{split}
        \theta^*, \phi^* &= \arg\max_{\theta, \phi} \mathcal{L}(\theta, \phi) \\
        \mathcal{L}(\theta, \phi) &= \mathcal{L}^{mle}(\theta) + \lambda \mathcal{L}^{vae}(\theta, \phi) \\
        &=\sum_{z \in B} \log P_\theta(z|d, q) + \lambda E_{P_\theta(z|d, q)} \log \frac{P_\phi(q|d, z)}{P_\theta(z|d, q)} \\
    \end{split}
\end{equation}
where $\mathcal{L}^{mle}(\theta)$ is the total log likelihood of the set of model-predicted solutions (denoted by $B$) with correct answer. $\mathcal{L}^{vae}(\theta, \phi)$ is the evidence lower bound of the log likelihood of questions. $\lambda$ is the coefficient of $\mathcal{L}^{vae}(\theta, \phi)$. The optimization process is divided into three stages: (1) the 1\textsuperscript{st} stage pre-trains a task-specific model $\theta$ with HardEM-thres on solution sets\footnote{\citet{DBLP:conf/conll/0001L18} pre-trained the task-specific model $\theta$ by maximizing $\mathcal{L}^{mle}(\theta)$. We enhanced their method by pre-training $\theta$ with HardEM-thres.}; (2) the 2\textsuperscript{nd} stage pairs the task-specific model with our question reconstructor $\phi$ to optimize $\mathcal{L}(\theta, \phi)$ for one epoch, except that $\mathcal{L}^{vae}(\theta, \phi)$ is used to pre-train $\phi$ and is kept from back-propagating to $\theta$; (3) the 3\textsuperscript{rd} stage optimizes $\mathcal{L}(\theta, \phi)$ while allowing $\mathcal{L}^{vae}(\theta, \phi)$ to back-propagate to $\theta$. The gradient of $\mathcal{L}^{vae}(\theta, \phi)$ w.r.t. $\theta$ is given by:
\begin{equation}
    \small
    \notag
    \begin{split}
        \bigtriangledown_\theta \mathcal{L}^{vae}(\theta, \phi) &= E_{P_\theta(z|d, q)} R \bigtriangledown_\theta \log P_\theta(z|d, q) \\
        R &= \log \frac{P_\phi(q|d, z)}{P_\theta(z|d, q)} \\
    \end{split}
\end{equation}
where $R$ is the reward function. To stabilize training, we use the average reward of 5 sampled solutions as a baseline $b$ and re-define the reward function as $R^{'} = R - b$. $\lambda$ is set to 0.1. 

In section 4.3, we report performance of the best model in the 3\textsuperscript{rd} stage. At the 2\textsuperscript{nd} stage, as the task-specific model optimized on both correct solutions and spurious solutions equally, the F1 score dropped from 72.35 to 67.93 at the end of this stage, indicating that correct training solutions is vital for generalization. At the 3\textsuperscript{rd} stage, model learning was further regularized with $\mathcal{L}^{vae}(\theta, \phi)$ which was optimized via reinforcement learning. Despite our efforts to stabilize training, the F1 score still dropped to 36.28 after training for 1,000 steps at the 3\textsuperscript{rd} stage.

\noindent
\textbf{Ours}: Suppose we have access to such an optimal question reconstructor $P_{\phi^*}(q|d, z)$ that $\mathbb{I}(f(z) = a)P_{\phi^*}(q|d, z)$ approximates $\log P_\theta(q, a|d, z)$ well at each training cycle, according to Eq. \ref{ori_obj}, after a sufficient number of training cycles, $\arg\max_{z \in Z} P_\theta(z|d, q, a)$ is expected to choose the same solution as $\arg\max_{z \in Z} P_{\phi^*}(q|d, z)$ does, otherwise the mutual information may not be maximal. In other words, a task-specific model can learn to choose solutions that are relevant to the questions if optimized on the mutual information maximization objective for sufficient steps, after which the question reconstructor is no longer needed. In fact, on WikiSQL which also provides ground-truth solutions, we observed that $\arg\max_{z \in Z} P_\theta(z|d, q, a)$ could choose solutions of better quality than $\arg\max_{z \in Z} P_\phi(q|d, z)$ did after sufficient training on our objective. We conjecture that this is due to approximation errors of the question reconstructor. Therefore, we switched to HardEM (without annealing) after optimizing our objective for a pre-specified number of steps which is tuned based on model performance on the development set.

\subsection{Experimental Settings}
For all experiments, we used previously proposed task-specific models and optimized them with their original optimizer. We chose the best task-specific model according to its performance on the development set.
As for our learning method, we used BART\textsubscript{base} as the question reconstructor. AdamW optimizer \cite{DBLP:conf/iclr/LoshchilovH19} was used to update the question reconstructor with learning rate set to 5e-5.

\subsubsection{Multi-mention Reading Comprehension}
We adopted the reading comprehension model, data preprocessing, and training configurations from \citet{DBLP:conf/emnlp/MinCHZ19}. 

\noindent
\textbf{Task-specific model}:
The model is based on uncased version of BERT\textsubscript{base}, which takes as input the concatenation of a question and a paragraph, and outputs the probability distribution of the start and end position of the answer span. To deal with multi-paragraph reading comprehension, it also trains a paragraph selector; during inference, it outputs a span from the paragraph ranked 1\textsuperscript{st}.

\noindent
\textbf{Data Preprocessing}:
Documents are split to segments up to 300 tokens. For Quasar-T, as retrieved sentences are short, we concatenated all sentences into one document in decreasing order of retrieval score (i.e., relevance with the question); for WebQuestions, we concatenated 5 retrieved paragraphs into one document, resulting in 10 reference documents per question.

\noindent
\textbf{Training}: Batch size is 20. BertAdam optimizer was used to update the reading comprehension model with learning rate set to 5e-5. The number of training epochs is 10.
We switched to HardEM after optimizing our objective for 10k steps and 7k steps on Quasar-T and WebQuestions, respectively.

\subsubsection{Discrete Reasoning over Paragraphs}
We used NeRd \cite{DBLP:conf/iclr/ChenLYZSL20} for discrete reasoning. The major differences with its original implementation have been discussed in section 4.3.

\noindent
\textbf{Task-specific Model}:
\citet{DBLP:conf/iclr/ChenLYZSL20} have designed a domain-specific language for discrete reasoning on DROP. The definitions of solutions for discrete reasoning introduced in section 4.3 are also expressed in this language except that we use different symbols (e.g., the minus sign \textit{``-''} in our definitions has the same meaning as the symbol \textit{``DIFF''} in their paper). NeRd is a Seq2Seq model which tasks as input the concatenation of a question and a paragraph, and generates the solution as a sequence. The answer is obtained by executing the solution.

\noindent
\textbf{Data Preprocessing}:
The input of the task-specific model is truncated to up to 512 words. We used the solution sets provided by \citet{DBLP:conf/iclr/ChenLYZSL20}, which cover 93.2\% of examples in the train set.

\noindent
\textbf{Training}:
Batch size is 32. Adam optimizer \cite{DBLP:journals/corr/KingmaB14} was used to update NeRd with learning rate set to 5e-5. The number of training epochs is 20.
We switched to HardEM after optimizing our objective for 1 epoch.

\subsubsection{Semantic Parsing}
Following \citet{DBLP:conf/emnlp/MinCHZ19}, we used SQLova \cite{DBLP:journals/corr/abs-1902-01069} on WikiSQL.

\noindent
\textbf{Task-specific Model}:
SQLova encodes the concatenation of a question and a table header with uncased BERT\textsubscript{base}, and outputs a SQL query via slot filling with an NL2SQL (natural language to SQL) layer.

\noindent
\textbf{Data Preprocessing}: Data preprocessing was kept the same as in \cite{DBLP:conf/emnlp/MinCHZ19}. We also used the solution sets provided by \citet{DBLP:conf/emnlp/MinCHZ19} which cover 98.8\% of examples in the train set.

\noindent
\textbf{Training}:
Following \citet{DBLP:conf/emnlp/MinCHZ19}, we set the batch size to 10. Following \citet{DBLP:journals/corr/abs-1902-01069}, Adam optimizer was used to update SQLova with learning rate of BERT\textsubscript{base} and NL2SQL layer set to 1e-5 and 1e-3, respectively. The number of training epochs is 15 and 20 when using the original train set and the hard train set of WikiSQL, respectively.
We switched to HardEM after optimizing our objective for 1 epoch and 9 epochs on the original train set and the hard train set, respectively.

\subsection{Computing Infrastructure}
We conducted experiments on 24GB Quadro RTX 6000 GPUs. Most experiments used 1 GPU except that experiments on DROP used 4 GPUs in parallel.

\section{Details of Ablation Study}
\subsection{SQL Selection Task}
We defined a SQL selection task on the development set of WikiSQL. Specifically, for each question, we randomly sampled $min(10, |Z|)$ solution candidates from the solution set $Z$ without replacement while ensuring the ground-truth solution was one of the candidates. A model was required to pick out the ground-truth solution by selecting the candidate with the highest prediction probability.

In section 5.3, we only show model accuracy in the first 10 training epochs because for \textit{BART\textsubscript{base} w/ HardEM}, \textit{SQLova w/ HardEM}, and \textit{SQLova w/ Ours}, model confidence (computed as the average log likelihood of selected SQLs) showed a downward trend after the 2\textsuperscript{nd}, 4\textsuperscript{th}, and $\ge$ 10\textsuperscript{th} epoch, respectively.

\subsection{Choice of Question Reconstructor}
We investigated how the choice of the question reconstructor affects results. One alternative choice is a Transformer pre-trained as a denoising auto-encoder of questions on the train set. This question reconstructor is the same as BART\textsubscript{base} except that the number of encoder layers and the number of decoder layers are 3 respectively. We pre-trained the question reconstructor for one epoch to reconstruct original questions from corrupted ones. For 50\% of the time, the input question is the original question; otherwise, we followed \citet{DBLP:conf/acl/LewisLGGMLSZ20} to corrupt the original question by randomly masking a number of text spans with span lengths drawn from a Poisson distribution ($\lambda$ = 3). Batch size is 4. AdamW optimizer was used with learning rate set to 5e-5.

\end{document}